\documentclass[preprint,review,12pt]{elsarticle}

\usepackage{graphicx} 
\usepackage{amsmath} 
\usepackage{amssymb}  
\usepackage{algorithm}
\usepackage{algorithmic}
\usepackage{dsfont}
\usepackage{booktabs}
\usepackage{multirow}
\usepackage{color}
\usepackage{siunitx}

\usepackage{verbatim}
\usepackage{float}
\usepackage{wrapfig}
\usepackage{subfig}

\usepackage{hyperref}
\hypersetup{
    colorlinks=true,
    linkcolor=blue,
    filecolor=magenta,      
    urlcolor=magenta,
}
 
\journal{Computer Vision and Image Understanding}

\begin{document}

\begin{frontmatter}



\title{Robust Face Tracking using Multiple Appearance Models and Graph Relational Learning}

\author[First]{Tanushri Chakravorty\corref{cor1}}
\ead{tanushri.chakravorty@polymtl.ca}

\author[First]{Guillaume-Alexandre Bilodeau}
\ead{gabilodeau@polymtl.ca}

\author[Third]{Eric Granger}
\ead{eric.granger@etsmtl.ca}

\cortext[cor1]{Corresponding Author}

\address[First]{LITIV Lab., Polytechnique Montr\'{e}al, Canada}
\address[Third]{LIVIA, \'{E}cole de technologie sup\'{e}rieure, Montr\'{e}al, Canada}

\begin{abstract}
This paper addresses the problem of appearance matching across different challenges while doing visual face tracking in real-world scenarios. In this paper, FaceTrack is proposed that utilizes multiple appearance models with its long-term and short-term appearance memory for efficient face tracking. It demonstrates robustness to deformation, in-plane and out-of-plane rotation, scale, distractors and background clutter. It capitalizes on the advantages of the tracking-by-detection, by using a face detector that tackles drastic scale appearance change of a face. The detector also helps to reinitialize FaceTrack during drift. A weighted \textit{score-level fusion} strategy is proposed to obtain the face tracking output having the highest fusion score by generating candidates around possible face locations. The tracker showcases impressive performance when initiated automatically by outperforming many state-of-the-art trackers, except Struck by a very minute margin: $0.001$ in precision and $0.017$ in success respectively.

\end{abstract}


\end{frontmatter}


\section{Introduction}\label{Introduction}
Face tracking has been studied for decades and it is still one of the challenging problems in computer vision. Face tracking in unconstrained videos promises to augment a wide range of applications in robotic vision, video analysis and face recognition, and is not only limited to visual surveillance. It is often used in video conferencing, but it is also useful in video-based face recognition as shown in \cite{saman}. It is defined as the task of locating a face in a given frame whether it is occluded or not. The face tracker is initiated in two ways: (1) using a ground-truth bounding box containing a face, (2) using a bounding box provided by a face detector. This box is also called an ROI (Region Of Interest). The output of the face tracker is the location of a face in a frame and is represented by a bounding box. 

As the face tracker outputs ROIs over a series of consecutive frames in a video sequence, it accumulates multiple evidence for the presence of a target face. Hence, the face tracker can preserve the identity of a target face since it works on the principle of \textit{spatio-temporal} information between consecutive frames. In contrast, a face detector searches for a face in the entire image, without any \textit{spatio-temporal} information, and thus cannot keep the identity of a face. 

Our primary contribution is to represent a face in a L2-subspace with a \textit{relational} graph. The term \textit{relational} describes the relation of features with the center of the bounding box during tracking initialization. This information comprises of three components: L2 distance of a feature with the center ($FDL$), importance of the feature ($w$), and feature descriptor ($D$). This model not only describes the appearance of the target face by representing it in a L2-subspace, but also encapsulates semantic information specific to the target face for occlusion. Thus, when this relational graph is discovered in a subsequent frame by matching feature descriptors, each matched feature outputs a center location of the target face using its L2-subspace representation. This center prediction is approximated by using multiple kernels in a response map reflecting the importance of each matched feature for the center prediction. The face localization is done by first concatenating all the generated kernel responses and then analyzing the peak response in the kernel map, which is transformed back to the cartesian coordinate system as face center location. Analyzing the peak in the map helps in eradicating the influence of errors during face localization, since multiple overlapped responses indicate reliable face center prediction over responses generated by tracking errors.

\begin{figure}[!htbp]
\vspace{-5em}
\centering
\includegraphics[width=\linewidth]{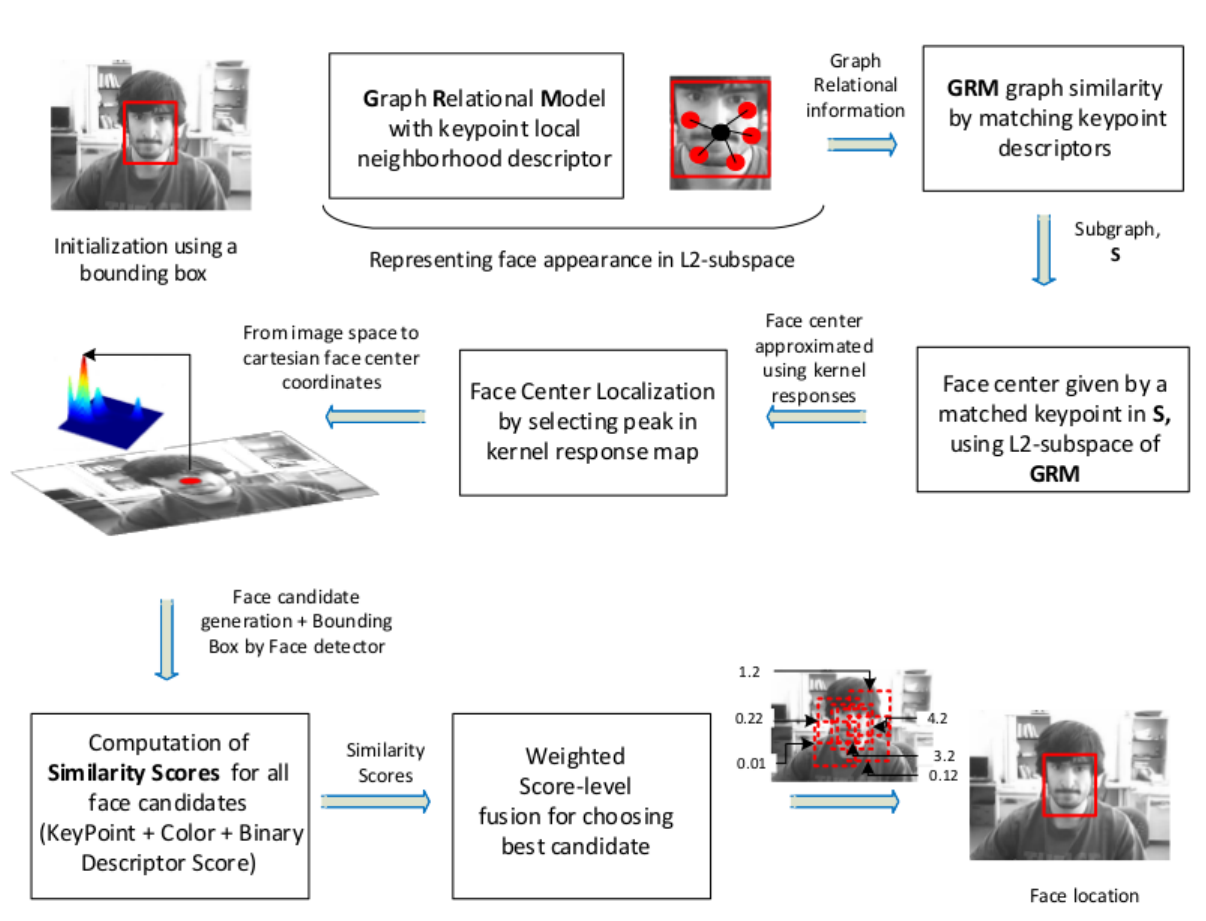}
\caption{Face localization process using Graph Relational Model.
 \label{fig:fig1}}
 \end{figure}

The relational graph is learned incrementally by adding and deleting connections in the graph during the \textit{appearance} model update. Since, the good connections are retained in the graph to help in localizing the center of the target face, this appearance model acts like a long-term memory of the target face. This appearance model is coined as GRM (Graph Relational Model), and is one of the proposed appearance models used in FaceTrack. The graph matching and face localization concept using GRM is illustrated in Figure \ref{fig:fig1}.

In contrast, the other proposed appearance models, ICM (Isotropic Color Model) and BDM (Binary Descriptor Model), help to find the target face during drastic appearance changes like illumination variation, in-plane rotation, out-of-plane rotation and heavy occlusion. The ICM describes the holistic face appearance, whereas the BDM helps to detect the intrinsic spatio-temporal changes happening at the pixel level. They both serve as a short-term memory of the current target face appearance, and are updated \textit{partially (and/or fully)}, depending on the occlusion detection strategy. By following this appearance model scheme for tracking, the \textit{temporal} information of a target face gets accumulated, and the tracker gets an appropriate appearance memory of the target face for appearance matching.

\begin{figure}[!htbp]
 \centering
\includegraphics[width=\linewidth]{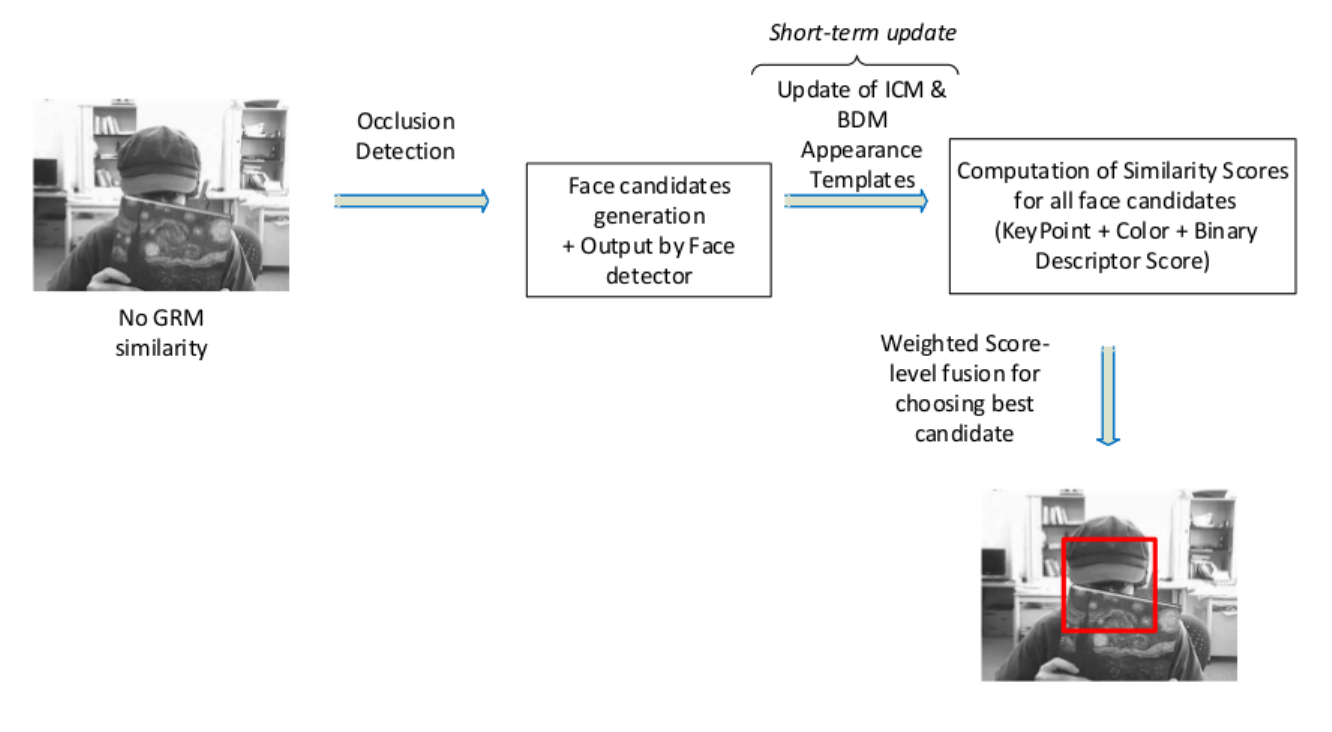}
\caption{Occlusion detection, tracking control and update strategy for the proposed face tracking system using GRM.
 \label{fig:occ}}
 \end{figure}

The GRM is effective as long as the graph structure remains visible and gets fully or partially matched. During other situations, the remaining appearance models (ICM \& BDM) are used for estimating the face location as shown in Figure \ref{fig:occ}. Apart from this, the appearance models are incrementally learned and the importance of features is determined on-the-fly for keeping a temporal memory, both long-term (GRM) and short-term (ICM and BDM), of the appearance of the target face. The proposed model is built to handle many tracking challenges like motion blur, fast motion, partial and heavy occlusion, background clutter and scale change. Each component plays a vital role in localizing the target face and the proposed tracker utilizes all the advantages from these components for accurate tracking. 

Our secondary contributions are a robust tracking strategy that assigns importance to appearance features during tracking \textit{initialization} and continues during the \textit{entire} face tracking process. The robustness is integrated using \textit{isotropy} to the appearance features used in tracking. The isotropic nature of features is formulated in a manner such that the feature closest to the center obtains the highest \textit{importance} as compared to others. By doing this, the background features that may get encapsulated in the appearance model, have lesser contribution in the kernel response map for target face center localization. In addition, the importance of the features get \textit{adapted} online and the lesser important features are deleted from the graph and the newer ones are added during model update, following the same policy of using isotropy to establish the importance to newly added features.  

Apart from this, we use a \textit{tracking-by-detection} approach by employing a face detector, \cite{Liao:2016:FAU:2914182.2914310}, with FaceTrack. The face detector helps to handle scale and aspect ratio changes of the face, drift and may help in reinitialization of the tracker during severe appearance changes. But, using either a single or multiple appearance based tracker with a face detector alone cannot effectively solve the face tracking problem. This is because the face detector focuses only on appearance similarities and ignores the spatio-temporal information in images, due to which there are large fluctuations in detection scores between two consecutive frames. On the other hand, the tracker might lose the target face due to large appearance variations. Hence, the face detector output is also used in face localization, thus capitalizing on their respective strengths. 

However, due to tracking noise and face deformation the localized face may not be precise. Hence, face candidates are generated around the localized face region obtained using face appearance matching with the help of multiple appearance models. Thus, in the proposed method, face tracking is considered as a problem of accurately estimating the \textit{face candidate} having the highest fusion score in a given frame. Hence, to obtain the final tracking output, a weighted \textit{score-level fusion} criteria is formulated for selecting the best face candidate.

\subsection{Contributions}
The main contributions of this paper are as follows:
\begin{enumerate}
\item A novel face tracking method is proposed that utilizes multiple appearance models to account for the temporal appearance matching of a target face for robust tracking.
\item A long-term and short-term strategy is proposed for effective matching during face tracking in real-world unconstrained video sequences.  
\item Robustness to face appearance features is integrated using isotropic weights. This ensures to obtain face localization using  importance face appearance features during the entire tracking process, thus tackling drift and background clutter.
\item A weighted score-level fusion approach is proposed for estimating the best face candidate as face location. 
\item A novel tracking control and update strategy that accounts for occlusion detection, tracking robustness and stability is proposed.
\end{enumerate}

The rest of the paper is organized as follows: Section \ref{sec:Rel} presents some related works in visual object tracking. Section \ref{sec:Method} discusses the proposed tracking framework in detail. Section \ref{sec:Experiments} provides the details of quantitative and qualitative experiments and analysis of each of the tracking method. Finally, Section \ref{sec:Con} concludes this paper.

\section{Related Work}\label{sec:Rel}
 
In this section, we focus on the visual object tracking works related to the class of discriminative appearance-based trackers. These discriminative appearance-based trackers behave like binary classifiers and distinguish the target object from the background. These discriminative trackers incorporate some form of model update during the visual tracking process and the classifier learns from samples online [\cite{TLD}, \cite{boosting}, \cite{Struck}, \cite{MIL}].

The TLD \cite{TLD} method uses a binary feature detector and an optical flow tracker. The detector learns from the examples which are sampled online from the bounding box. Positive examples are labeled from the region inside the box and the negative examples are taken from the region around the bounding box. In contrast, MIL \cite{MIL} utilizes Haar features as samples which are grouped into a bag. Along with the bounding box, the tracker uses rectangular windows around the nearby region as positive samples, since the target region can include some background region. Negative bags comprise of rectangular boxes which are farther from the bounding box. In Boosting \cite{boosting}, the method employs a boosting classifier based on Haar features for selecting discriminative features for distinguishing the target object from the background. In Struck \cite{Struck}, Haar features from the box are considered as an appearance model for tracking. In their method, instead of generating samples from around the bounding box, the samples are generated by translating the bounding box and then fed to a SVM (Support Vector Machine) classifier. Thus, the sampling strategy for Struck is different from the aforementioned tracking methods. However, the classifier learning is constrained by maintaining a budget that helps to maintain a set of the support vectors. Recently, correlation filter learning method like \cite{KCF}, has shown impressive results due to its dense feature extraction and sampling technique for high-speed tracking.

Detections provided by object detector are used in tracking objects whose prior information is known. The trackers of \cite{TLD} and \cite{boosting} are special cases of tracking-by-detection. The detections enable the tracking process to tackle scale appearance change and sometimes drift. Similarly, a face detector is used in FaceTrack to tackle drastic appearance change such as scale change between two consecutive frames and reinitializes it during drift. 
 
Ross et al. introduced incremental subspace learning in visual object tracking with the concept that the target can be represented in a low dimensional subspace that can be helpful in dealing with tracking nuisances, like pose and illumination variation \cite{IVT}. This idea works well in situations where the errors are small and localized, i.e., they follow a Gaussian distribution. However, in some scenarios like when there is occlusion, the errors might be large. In such cases, this type of global representation might not be able to cope up and thus result in track loss. To overcome this, the authors in \cite{6619151}, assumed that tracking errors follow a Gaussian-Laplacian distribution. Owing to their success, their error-removing method is employed in various works \cite{6926840}, \cite{MIL}. In real-world scenario the data can however, contain various types of noise, and the data or noisy samples that may belong to other targets may get included in the appearance model of the target and ultimately degrade the performance, particularly for graph-based learning methods. Hence, authors in \cite{Ehsan} proposed a spectral clustering method, which consider edges with higher weights in the graph cluster and segment other parts in the graph.

In our approach, the GRM model adds new samples to the relational graph by taking keypoints from the tracking bounding box itself, thus removing the need for using segmentation and clustering.

Besides this, adaptive appearance models like \cite{aam}, \cite{Dewan}, use face tracking for face recognition purpose. They use online samples for updating the appearance of the face, and employ forgetting factor for adapting the appearance model.
Related to our work are object trackers  \cite{TUNA}, \cite{CTSE} and \cite{CMT} that utilize structure of the object as the appearance representation for tracking.

In contrast to these approaches, our method maintains a temporal appearance memory using multiple appearance models of the target face that leverages the benefit of both long-term and short-term appearance updates that are proven essential for robust face tracking. Moreover, we adapt the face appearance representation such that potentially distracting regions are suppressed in advance and thus, no explicit tracking of distractors (similar looking faces) is required, thus ensuring stable face tracking in real-world scenarios. The next section details the workings of FaceTrack.

\section{Proposed Face Tracking Method}\label{sec:Method}
It has been shown in \cite{CTSE} and \cite{CMT} that structure can be a powerful appearance representation for visual object tracking. Whereas in \cite{TUNA}, it has been shown that the structure of an object can help to tackle occlusion. Our motivation for using structure is inspired by the idea that by exploring the \textit{intrinsic} structure of a target face may help to discover a particular pattern of a face of interest. 

In machine learning tasks such as subspace learning, semi-supervised learning and data clustering, informative directed or undirected graphs are used to study the pairwise relationships between data samples that helps to identify a pattern belonging to a specific object \cite{Shi:2000:NCI:351581.351611}. Thus, for identifying a particular pattern belonging to a face of interest, a graph has the following characteristics that can be highly beneficial for face tracking:
\begin{itemize}
\item Distinct Representation: Graphs are powerful representation tools. Higher dimensional data can be represented in a manner which can be utilized for problem solving. 
\item Relational information: Graphs can help to identify the internal structure which can be utilized by relational information such as metric between points in the graph, rather than just the attributes of the entities being present \cite{Holder:2003:GRL:959242.959254}.
\item Sparsity: Findings in subspace learning \cite{Belkin:2003:LED:795523.795528} show that sparse graph characterizes local relations and thus can help in better classification.
\end{itemize}
Hence, the aforementioned advantages of a graph can be used for building a robust appearance model for face tracking in videos.

\begin{table}[!htbp]
\centering
\begin{tabular}{p{0.15\textwidth}p{0.1\textwidth} p{0.7\textwidth}}
\toprule
{\textit{Appearance Model}}  & {\textit{Notation}} & \hspace{1mm}Feature Description\\
\midrule
Graph Relational Model, & $D$ & SIFT keypoint features at each location and scale in an image, thus are multi-scale and spatially specific with their invariant keypoint descriptor \cite{Lowe:1999:ORL:850924.851523}.\\
{\textbf{GRM}}& $FDL$ & Represents face appearance in L2-subspace by encoding \textit{L2 distance} of a SIFT keypoint from the tracked bounding box center, denoted as $FDL=[\Delta x, \Delta y]$.\\
& $w$ & Describes the importance of a keypoint assigned using \textit{isotropy}.\\
\\
Isotropic Color Model, {\textbf{ICM}} & & Holistic discriminative feature of the face (tracked bounding box), 3-channel \textit{Gaussian} weighted color histogram (pixels are assigned importance using \textit{isotropy}.) \\
\\
Binary Descriptor Model, {\textbf{BDM}} & & Encodes spatio-temporal local neighborhood information of a pixel into a 1-channel, 16-bit LBSP binary descriptor, \cite{lbsp}.\\
\bottomrule
\end{tabular}
\caption{Multiple Appearance Models used in FaceTrack}
\label{tab:Mult}
\end{table}

\subsection{Building the multiple appearance models}
\label{Building the multiple appearance models}

The proposed model characterizes the target face contained in the initialized bounding box by using multiple appearance models namely GRM, ICM and BDM respectively. GRM characterizes the face from two perspectives. First, by encoding features related to a face by detecting and describing keypoint descriptors that belong to the face. Second, is by representing the keypoints in the L2 subspace by forming a relation between the detected keypoints with the center of the initialized box using relational information. It is robust towards partial occlusions and deformations, as a visible part of the GRM can still output the target face center by using its relational information. For more robustness, during the model initialization, the keypoints that are closer to the center are given higher \textit{importance} and are assigned higher \textit{weights} as compared to others that are farther from the center. The weight associated to the importance is given by Equation \ref{eq:weight}:

\begin{equation}\label{eq:weight}
w_{k_{i}} = \max((1 - |\eta\cdot FDL|), 0.5);
\end{equation}

The relational information for a keypoint in GRM is represented as $\{FDL,D,w\}$, where \textit{FDL} is the L2 subspace representation of a feature point with the graph center, $D$ is the keypoint descriptor, and $w$ is the weight (importance) of a feature. Furthermore, ICM encodes the holistic appearance using color histogram, and BDM encodes the spatio-temporal neighborhood local information for the pixels contained in the initialized bounding box. Table \ref{tab:Mult} summarizes the multiple appearance models with their respective feature description.

\subsection{Graph similarity matching using GRM}\label{subsec:grm}

With every new frame being processed by detecting and describing keypoints, our method tries to find a subgraph, $S$, in the frame that maximizes the similarity with the GRM by matching their keypoint descriptors. Let us denote an object GRM as $G$. For finding the center of the face, we will use the L2-subspace representation, $FDL$ and its importance, $w$, associated with the matched keypoints to obtain the center. Hence, we are trying to find a subgraph, $S$ which is \textit{isomorphic} to $G$ or to a subgraph of $G$. Thus, the maximum similarity between the two graphs can be represented as a function given by Equation \ref{eq:sim}:
\begin{equation}\label{eq:sim}
sim(G,S) = D(G)\sqcap_{m}D(S)
\end{equation}

where, $\sqcap_{m}$ is the \textit{bijection} that represents the keypoint matches, and $D$ is the feature descriptors of the two graphs respectively. The total number of matched keypoint descriptors with $D$, at current frame, $t$, is given by $N$\footnote{\cite{Lowe:1999:ORL:850924.851523} uses ratio test to eradicate matches higher than $0.8$. In FaceTrack experiments $0.75$ is used.}. Now, we use this similarity knowledge to get the face center by using the relational information $FDL$ (L2-subspace), and matched keypoint, $k$ in $S$. Thus, the face center given by a matched keypoints in $S$, can be represented using Equation \ref{eq:simK}.
\begin{equation}\label{eq:simK}
x^{t}_{{Center}^{k}}=  k_{x,y} + FDL
\end{equation}

However, the subgraph may contain errors due to noise. Further, their structure cannot be determined in advance. Therefore, $x^{t}_{{Center}^{k}}$ is approximated using kernel responses denoted as $\varphi$. Two kernel functions are used for generating the response: Gaussian kernel,  $\Phi_{1}$, and Exponential kernel,  $\Phi_{2}$, respectively\footnote{The Gaussian kernel parameters are $\sigma=6.0$, with a $5\times5$ filter size. The denominator, $\Theta$ of the Exponential kernel is taken as $8000.0$.}, and $\varphi_{k}$ is represented by Equation \ref{eq:OptKernel}:
\begin{equation}\label{eq:OptKernel}
\varphi_{k}= \Phi_{1}(x^{t}_{{Center}^{k}}).\Phi_{2}(x^{t}_{{Center}^{k}} - x^{t-1}_{Center}).w
\end{equation}
where, $w$ is the importance of a matched feature keypoint in $G$ and $x^{t-1}_{Center}$ is the face location in frame $t-1$. Now, all the $N$ kernel responses are accumulated, i.e, they get overlapped. The face center location is obtained by analyzing the peak in the kernel response map, and is given by Equation \ref{eq:mle}:
\begin{equation}\label{eq:mle}
x_{Center}^{t} = \underset{x}{\max}\left ( \sum_{k=1}^{N}\varphi_{k}(x) \right )
\end{equation}

The obtained peak response is transformed back into the image coordinate system to obtain the face center location. As shown in Figure \ref{fig:fig1}, the peak (color coded as dark red) corresponds to the face center location. Hence, $x^{t}_{Center}$ denotes the optimal solution for the face center target obtained by GRM model at frame $t$. While analyzing the kernel map, it is noted that the response is \textit{anisotropic}, because of the different overlapping rates of the individual responses in the kernel map. This type of response proves highly beneficial for face localization by GRM from a regression perspective. In our method, during the kernel response generation, $\Phi_{1}$ is centered at the face center location given by Equation \ref{eq:simK}, such that it gets the highest value. On the other hand, $\Phi_{2}$ is highest when the face center given by the matched feature using Equation \ref{eq:simK}, is closer to the peak, $x^t_{Center}$. This helps to gain leverage over the short-term matched features in GRM that become relevant in generating kernel responses. As seen later in subsection \ref{subsec:expo}, by analyzing the response for the features that are outputting correctly for the center, their influence in the kernel response map increases and reduces for others that are predicting wrongly or farther from the $x^t_{Center}$.

\subsection{Scale Adaptation and Computation of Appearance Similarity Scores}

To adapt to the scale variation of the face, we use the same strategy, as used in \cite{TUNA}. The authors utilized pairwise distances between matched keypoints between consecutive frames to tackle scale change. Now for the output face location obtained using GRM, denoted as $x^{t}_{Center}$, face \textit{candidates} are generated around it, to improve localization precision, since the center may get shifted due to face deformation or tracking noise. Apart from this, the second component of the framework, i.e. the face detector, outputs a bounding box for a detected face for frame $t$. The obtained bounding box from the detector is also considered as a face candidate.

\begin{table}[!htbp]
\centering
\begin{tabular}{p{0.35\textwidth} p{0.55\textwidth}}
\toprule
{\textit{Similarity Scores}}  & Description \\
\midrule
Keypoint Score & 
$K_{fc_{i}} = \frac{n}{N}$, \\
\\
Color Score & $C_{fc_{i}} = \sqrt{\sum_{i=1}^{d}(ICM_{am}-ICM_{fc_{i}})^{2}}$, \\
\\
Binary Descriptor Score & $B_{fc_{i}} = BDM_{am}\oplus BDM_{fc_{i}}$,  \\
\bottomrule
\begin{tabular}{p{0.9\textwidth}}
$n$, is the number of matched keypoint descriptors present in face candidate, $fc$, \\
$N$, is total number of matched keypoint descriptors of GRM that were matched at frame $t$,\\
$d$, is the feature dimension,  $am$ denotes the appearance template model for ICM and BDM,\\
$\oplus$ represents an operation.
\end{tabular}
\\
\bottomrule
\end{tabular}
\caption{Computation of face appearance Similarity Scores in FaceTrack}
\label{tab:score}
\end{table}

\begin{algorithm}[!htbp]
\begin{algorithmic}[1]
  \caption{\textbf{FaceTrack Algorithm}
    \label{algo:tracking}}
    \FOR{all keypoints matched in subgraph, S}
    \STATE obtain face location using Equation \ref{eq:mle} and generate face candidates
    \STATE adapt scale using pairwise keypoint distance
            \FOR {all face candidates from GRM and face detector, at frame $t$} 
            \STATE compute Similarity Scores, refer Table \ref{tab:score}
            \STATE compute variance of Similarity Scores
            \STATE compute $FS_{fc_{i}}$ using Equation \ref{eq:rankScore}
            \ENDFOR
            \ENDFOR
            \STATE best face box as face candidate with max $FS_{fc_{i}}$
    \STATE update appearance models using Algorithm \ref{algo:trackingUpdate} 
\end{algorithmic}
\end{algorithm}

Next, for all the face candidates, the ICM and BDM models are first computed, and are matched for similarity. Table \ref{tab:score} describes the formula for the computation of the respective similarity scores. The ICM model is compared using the norm \textit{L2} norm, and is called Color Score, $C_{fc}$. The BDM model is compared using \textit{hamming} distance, and is called by Binary Descriptor Score, $B_{fc}$. Further, a Keypoint Score, $K_{fc}$, for the matched keypoints in GRM, lying inside inside a face candidate box is computed. The features are normalized and transformed to the range $[0,1]$. All the similarity scores, $K_{fc}$, $B_{fc}$, and $C_{fc}$, associated with $fc$, are used for obtaining the best face box by using a weighted score-level fusion strategy, as we will see later in subsection \ref{sec:rank}.

\subsection{Face Localization using Weighted Score-level Fusion Strategy}\label{sec:rank}
For choosing the best candidate as the final output by the face tracking framework, we propose a strategy that combines the fusion of all the similarity scores (refer Table \ref{tab:score}), with weights based on their variance between two consecutive frames, such that the similarity score having the largest variance, gets the largest weight. If we just take the similarity score into account without its weighted variance, the fusion score might get higher for a candidate (e.g. distractor), even though it is not the face of interest which is required to be tracked. Moreover, the information from each appearance model are uncorrelated and by following this strategy, the contributions from each component can be utilized for maximum similarity. Thus, the best face candidate should maximize the following Equation \ref{eq:rankScore}.
\begin{equation}\label{eq:rankScore}
FS_{fc_{i}} = p\cdot K_{fc_{i}} + q \cdot C_{fc_{i}} + r \cdot B_{fc_{i}}
\end{equation}

where, $p$, $q$ and $r$ represent the weights assigned to the similarity scores, based on their variance ranking. The weights are assigned such that if $var(K_{fc_{i}}) > var(C_{fc_{i}}) > var(B_{fc_{i}})$, then $p$ gets multiplied with $K_{fc_{i}}$, $q$ with $C_{fc_{i}}$, and $r$ with $B_{fc_{i}}$, respectively. The ranking helps to determine the dominant similarity score in a face candidate, and fusion helps to choose the best candidate that maximizes all the similarity scores. Algorithm \ref{algo:tracking} summarizes the proposed tracking framework.

\subsection{Occlusion Detection, Tracking Control and Update Strategy}\label{subsec:expo}
We consider two complementary aspects in tracking, robustness and stability, by long-term and short-term update. Long-term update are performed during the whole tracking duration for all the keypoint features, $k_{i}$, collected for GRM model at frame $t$, by adapting their weights using Equation \ref{eq:adapt}:
\begin{equation}\label{eq:adapt}
w_{k_{i}}^{t+1}= 
\begin{cases}
(1 - \tau)w_{k_{i}}^{t}  + \tau\cdot\theta(l), & \text{if $k_{i} \epsilon N$},\\
(1 - \tau)w_{k_{i}}^{t},              & \text{otherwise}
\end{cases}
\end{equation}

where $\tau$ is \textit{learning rate}. The value of $\theta(l)$ increases with a keypoint prediction closer to the $x^{t}_{Center}$ and is obtained using Equation \ref{eq:theta} as:

\begin{equation}\label{eq:theta}
 \theta(l) = \max((1 - |\eta\cdot l|), 0.0);
\end{equation}
where $l$ is the L2 distance between the center location given by the matched feature keypoint using its relational information, and the center obtained by analyzing the response in the kernel map, $x^{t}_{Center}$. On the other hand, tracking control is done by analyzing the center response given by a matched keypoint. It is done to avoid potential tracking failures. For example, for a given frame $t$, if a matched keypoint outputs a center farther from the center ($x^{t}_{{Center}^{k}}$) in frame $t-1$, then its influence in the kernel response map for future frames get reduced using exponential kernel function, $\Phi_{2}$ (used in Equation \ref{eq:OptKernel}). It is given by the following Equation \ref{eq:expoFunc}:
\begin{equation}\label{eq:expoFunc}
\Phi_{2}\propto \exp\frac{-(x^{t}_{{Center}^{k}} - x^{t-1}_{Center})}{\Theta}
\end{equation}

By \textit{controlling} this, potential tracking drift failures can be avoided, which in turn gives the proposed method stability along with its robustness towards face appearance changes.

When the similarity between graphs cannot be established in a frame (i.e. no subgraph $S$ can be matched), we consider this scenario as an occlusion detection, and perform short-term update by partially (or fully)\footnote{partial update: by \textit{replacing} 12.5\% of the face appearance features in ICM model and 10\% of the face appearance features in BDM model respectively, full update: by \textit{replacing} 100 \% of the ICM and BDM model} updating the ICM and BDM model respectively. 
\begin{algorithm}[!htbp]
\begin{algorithmic}[1]
  \caption{\textbf{FaceTrack occlusion detection, control \& update strategy}
    \label{algo:trackingUpdate}} 
    \FOR  {keypoints, $k_{i}$, in GRM}
    \STATE Long-term update using Equation \ref{eq:adapt}
    \STATE Tracking control using Equation \ref{eq:expoFunc}
    \IF {$w_{k_{i}}$ $< \gamma$}
    \STATE Remove $k_{i}$ from GRM
    \ENDIF
    \IF {(N == 0)}
    \STATE Occlusion detected
    \IF{(appearance templates size != best face box size)}
    \STATE{\textit{partial} update of ICM \& BDM models}
            \ELSE
    \STATE {\textit{full} update of ICM \& BDM models}        
            \ENDIF
           \ENDIF
            \ENDFOR
             \IF {$K_{fc_{i}}> \alpha$ and $B_{fc_{i}}> \beta$}
    \STATE Add new keypoints in GRM
    \STATE \textit{full} update of ICM \& BDM models        
    \ENDIF
\end{algorithmic}
\end{algorithm}

During this scenario, the ICM and BDM models help to localize the face target, since the similarity scores of these models will dominate for the best face candidate. New features are added to GRM when the ICM and BDM similarity matching score for the face output template is above $\alpha$ and $\beta$, respectively. Features having weights lower than $\gamma$, are removed from GRM. 

Thus, by following this control and update strategy, the different appearance models complement each other during different tracking scenarios. Algorithm \ref{algo:trackingUpdate} summarizes the update strategy of the proposed face tracking framework.

\begin{table}[t]
\centering
\scalebox{0.9}{
\begin{tabular}
{ccccccccccc}
\toprule
 Video Attributes & MB & FM & BC & DEF & IV & IPR & OCC & OPR & OV & SV \\ 
      \midrule
    Total Number & 5 & 5 & 4 & 4 & 5 & 12 & 7 & 13 & 1 & 10\\
    \bottomrule
  \end{tabular}}
\caption{Distribution of attributes of the 15 video sequences: Motion Blur (MB), Fast Motion (FM), Background Clutter (BC), Deformation (DEF), Illumination Variation (IV), In-plane Rotation (IPR), Occlusion (OCC), Out-of-plane-Rotation (OPR), Out-of-View (OV), Scale Variation (SV).}
\label{tab:seqDist}
\end{table}

\section{Experimental Evaluation}\label{sec:Experiments}
The proposed method is validated on OTB benchmark \cite{ootb} for One-Pass Evaluation (OPE). The selected state-of-the-art trackers used for comparison are: Struck \cite{Struck}, TLD \cite{TLD}, KCF\cite{KCF}, MIL \cite{MIL}, CMT \cite{CMT}, TUNA \cite{TUNA} and Boosting \cite{boosting}. 15 video sequences from the benchmark containing faces are chosen for evaluation. These video sequences display several challenges that are encountered during tracking a face in a video sequence: occlusion (OCC), fast motion (FM), illumination variation (IV), scale variation (SV), motion blur (MB), in-plane-rotation (IPR), out-of-plane rotation (OPR), background clutter (BC), out-of-view (OV) and deformation (DEF). Table \ref{tab:seqDist} shows the distributions of attributes for the 15 face video sequences with different challenges.

\subsection{Evaluation Metrics}
The benchmark is evaluated on two performance measures: precision and success. \textit{Precision} is measured as the distance between the centers of a bounding box outputted by the tracker and the corresponding ground truth bounding box. The precision plot shows the percentage of frames whose center localization output are within a given threshold distance. \textit{Success} is measured as the intersection over union of pixels bounding box outputted by the tracker with the ground truth bounding box. The success plot shows the percentage of frames with their overlap score higher than a set of all the given thresholds, $t$, such that $t$ $\epsilon$ $[0,1]$.

For our experiments, we test all the trackers by initializing them in two ways: (1) Ground truth initialization, (2) Automatic initialization using a face detector \cite{Liao:2016:FAU:2914182.2914310}\footnote{Any other face detector can be used for initialization purpose.}. All the selected trackers for comparison are implemented in the OpenCV 3.1.0 library except Struck\footnote{https://github.com/samhare/struck}, CMT\footnote{https://github.com/gnebehay/CppMT} and TUNA\footnote{https://github.com/sinbycos/TUNA}, for which the code is provided online by the authors. The trackers are evaluated using the default parameters provided in their respective research papers. The proposed FaceTrack is tested on machine with configuration as Intel Core i7 @ 3.40GHz, 16GB RAM and is implemented in C++.For evaluation, the parameters of FaceTrack are: $\alpha= 0.23$, $\beta=0.1$, $\gamma=0.1$, $p= 0.15$, $q= 0.1$, $r=0.1$, $\tau= 0.9$, and $\eta = 0.005$. They are fixed for all the experiments. Face tracking results can be found at \href{http://step.polymtl.ca/~Tanushri/FaceTrack/}{http://step.polymtl.ca/$\sim$Tanushri/FaceTrack/}.

\begin{figure*}[t]
\vspace{-5em}
\captionsetup[subfigure]{labelformat=empty}
\centering
\subfloat[(a)]{\includegraphics[width=0.5\textwidth]{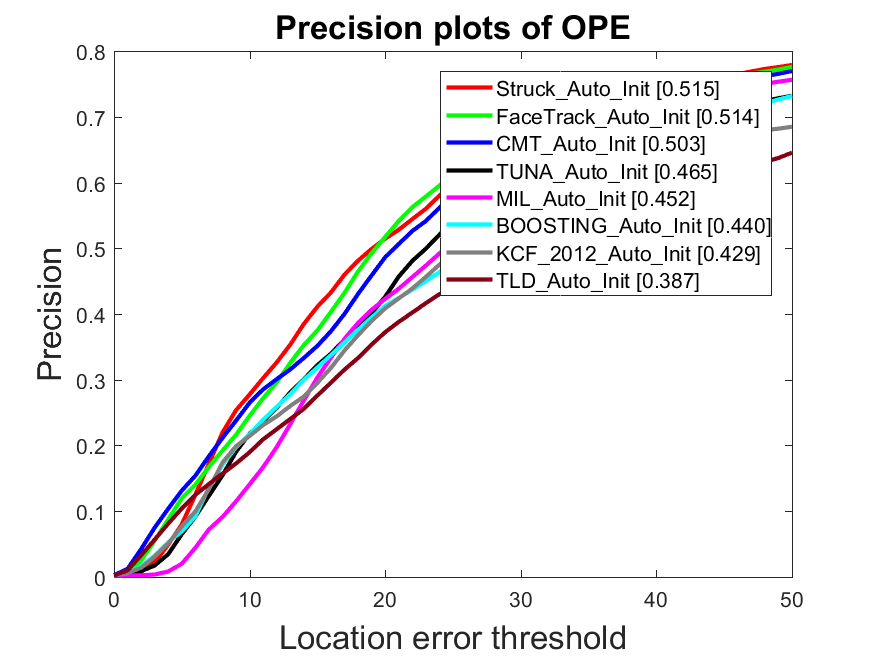}}
\subfloat[(b)]{\includegraphics[width=0.5\textwidth]{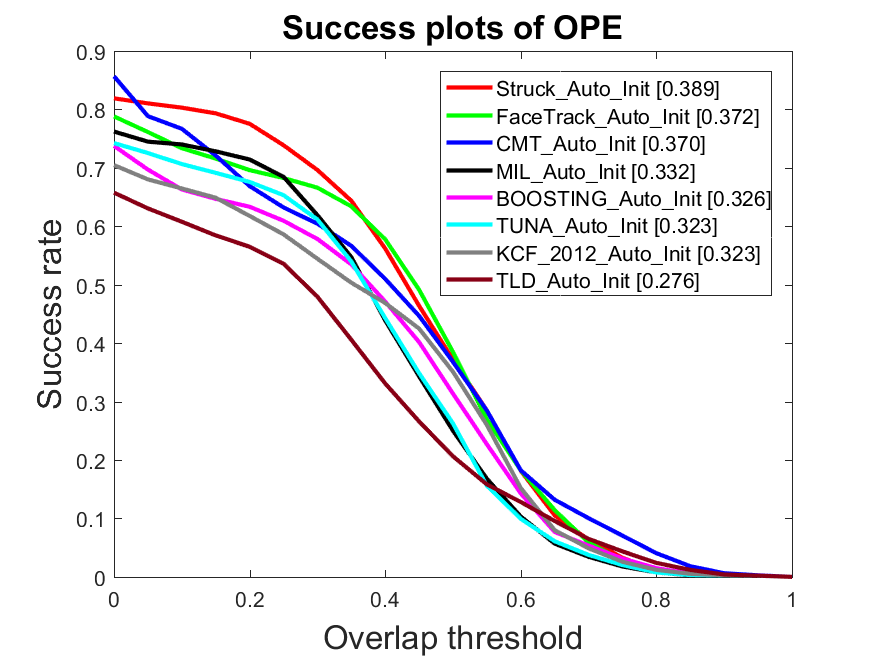}}
\caption{\footnotesize{FaceTrack performance in real-world scenarios when initialized automatically: (a) overall precision and (b) overall success. (Best viewed when zoomed in.)}}
\label{fig:AutoInitFT}
\end{figure*}

\subsection{Comparison to state-of-the-art}
FaceTrack shows strong performance when initialized using ground truth: precision, $0.603$ and success, $0.425$, respectively. Furthermore, it outperforms and ranks \textit{second} in overall performance when initialized automatically (Refer Figure \ref{fig:AutoInitFT}). This performance with automatic initilization showcases that FaceTrack is comparatively less affected by initialization. The robustness of FaceTrack can be attributed to its robust initialization strategy using \textit{isotropy} in which all features are not given equal importance. The keypoint features that get matched with GRM appearance model output a subgraph, containing an estimate of a region that contains the target face. However, for finer estimation for face location, the kernel responses of the matched features are summed. This response is \textit{anisotropic}, which efficiently determines precise face location, as it corresponds to maximum value of the cumulative overlapped responses. The response is guided through long-term update of features that are analyzed using multiple kernel functions (Refer subsections \ref{subsec:grm} and \ref{subsec:expo} for details). Further, for finer precision, face candidates are generated around this location and the bounding box given by the face detector is also considered a candidate. Finally, the weighted score-level fusion score helps to decide the best face candidate.

\begin{table}[!htbp]
\centering
\begin{tabular}
{p{0.17\textwidth}p{0.1\textwidth}p{0.1\textwidth}p{0.13\textwidth}p{0.1\textwidth}p{0.1\textwidth}p{0.13\textwidth}}
\toprule
{\textit{Algorithm}} &  &\textit{Precision}  & & & \textit{Success} \\
\midrule
& {\textit{GT Init}} & {\textit{Auto Init}} & {\textit{\%Relative change in Precision}}   & {\textit{GT Init}} & {\textit{Auto Init}} & {\textit{\%Relative change in Success}}\\
\midrule

\textbf{FaceTrack (Proposed)} & 0.603 & 0.514 & 14.76\% $\downarrow$ & 0.425 & 0.372 &  12.47\% $\downarrow$ \\
\\
Struck \cite{Struck} & 0.705 & 0.515 & \textbf{26.95}\% $\downarrow$ & 0.543 & 0.389 &  \textbf{28.36}\% $\downarrow$\\
\\
TLD \cite{TLD} & 0.432 & 0.387 & 10.42\%$\downarrow$ & 0.335 & 0.276 &  17.61\%$\downarrow$  \\
\\
KCF \cite{KCF} & 0.623 & 0.429 & \textbf{31.14}\%$\downarrow$ & 0.478 & 0.323 & \textbf{32.43}\%$\downarrow$ \\
\\
MIL \cite{MIL} & 0.496 & 0.452 &  8.87\% $\downarrow$ & 0.383 & 0.332 & 13.32\% $\downarrow$ \\
\\
Boosting \cite{boosting} & 0.520 & 0.440 &  15.38\% $\downarrow$ & 0.419 & 0.326 & 22.20\% $\downarrow$ \\
\\
CMT \cite{boosting} & 0.632 & 0.454 &  \textbf{28.16}\% $\downarrow$ & 0.502 & 0.333 & \textbf{33.67}\%$\downarrow$ \\
\\
TUNA \cite{TUNA} & 0.598 & 0.465 &  22.24\% $\downarrow$ & 0.475 & 0.323 & \textbf{32.00}\% $\downarrow$ \\
\bottomrule
\end{tabular}
\caption{Comparison of FaceTrack with the state-of-the-art trackers on 15 video sequences with various challenges. The bold text showcases the trackers most affected towards initialization.}
\label{tab:sum}
\end{table}

The uniqueness of GRM lies in its design as it helps in tracking a specific face. The approximation of $FDL$ using the Gaussian kernel helps to tackle face deformation, which happens very often during face tracking. During deformation, the keypoint feature can move by a pixel which can result in error. Thus, approximating the response using a Gaussian kernel compensates for this error, and in turn for face deformation. Even during heavy occlusion, in-plane rotation, the short-term updates help to locate the target face as the appearance matching can still be established with the aid of multiple appearance models during such scenarios. On the other hand, during drastic appearance changes like scale change, the face detector tackles it even if the some of the keypoint features in GRM may fail to get matched. However, in cases when no appearance matching can be established and the face detector also fails to detect a face, then the face location is not updated until the face appearance matching starts establishing again. However, it might be possible that the face detector outputs false positives. In addition, since it does not use any spatio-temporal information of the target face from the previous frame, its detection might be for a distractor. Therefore, in this case, the face candidates generated around the localized face by the GRM model will dominate in localizing face since their similarity score of appearance will be higher, thus, avoiding wrong face localization.

It is interesting to note that the performance results become more interesting when FaceTrack is initialized automatically and ranks just after Struck by a very minute margin. It can be noted in Table \ref{tab:sum} that the percentage drop in terms of performance is on the higher side, almost double for Struck, KCF, CMT and TUNA as compared to FaceTrack indicating that FaceTrack is less affected by the initialization as it gets re-initialized periodically when the face candidate sample is chosen. Moreover, the proposed occlusion detection, tracking control and update strategy that helps FaceTrack robust towards appearance changes but at the same time be less affected from distractions, thus outputting stable results. In addition, the use of the face detector aids in drastic appearance changes of target face. This also indicates that the model update which involves addition of new features and deletion of bad features that are not predicting for center in GRM, partial and full update of ICM and BDM models is most of the time happening correctly. An untimely update might result in corrupting the appearance models, and the tracker might fail. The next subsection gives detailed attribute-wise analysis of FaceTrack and how it is able to tackle various tracking challenges.

\subsection{Attribute-wise Analysis}
\begin{figure*}[!htbp]
\vspace{-10em}
\captionsetup[subfigure]{labelformat=empty}
\centering
\subfloat[(a)]{\includegraphics[width=0.5\linewidth]{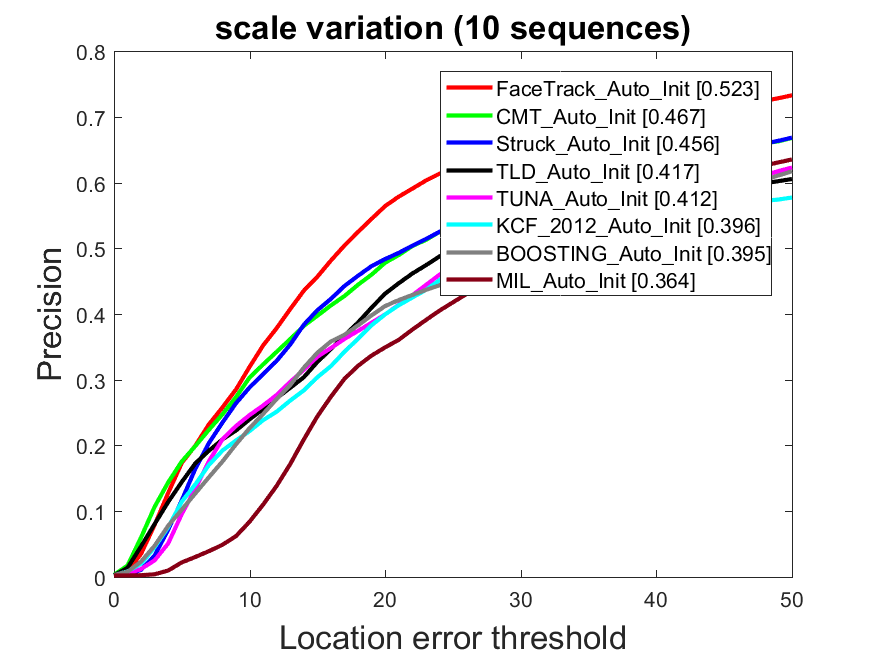}}
\subfloat[(b)]{\includegraphics[width=0.5\linewidth]{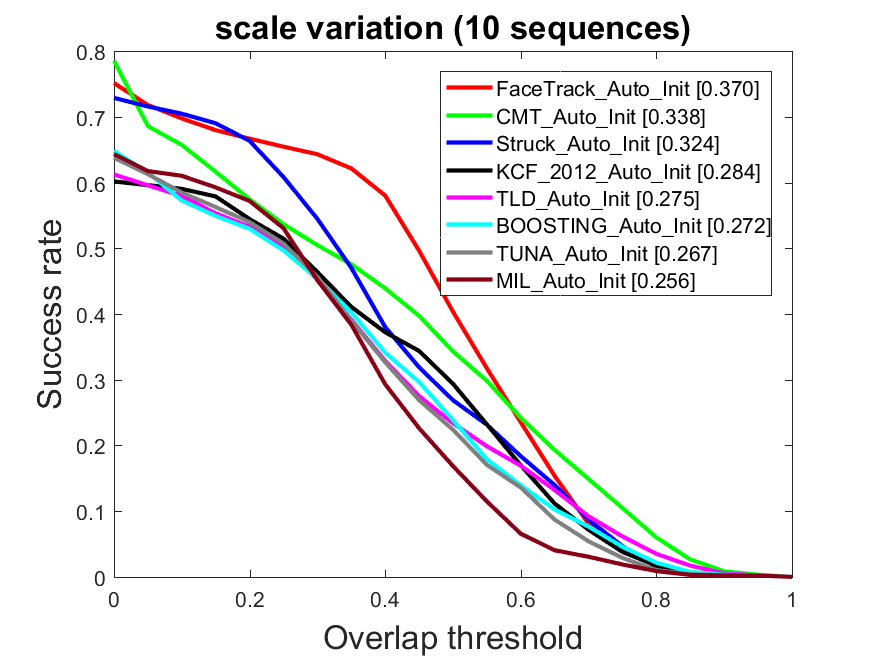}}\\
\subfloat[(c)]{\includegraphics[width=0.5\linewidth]{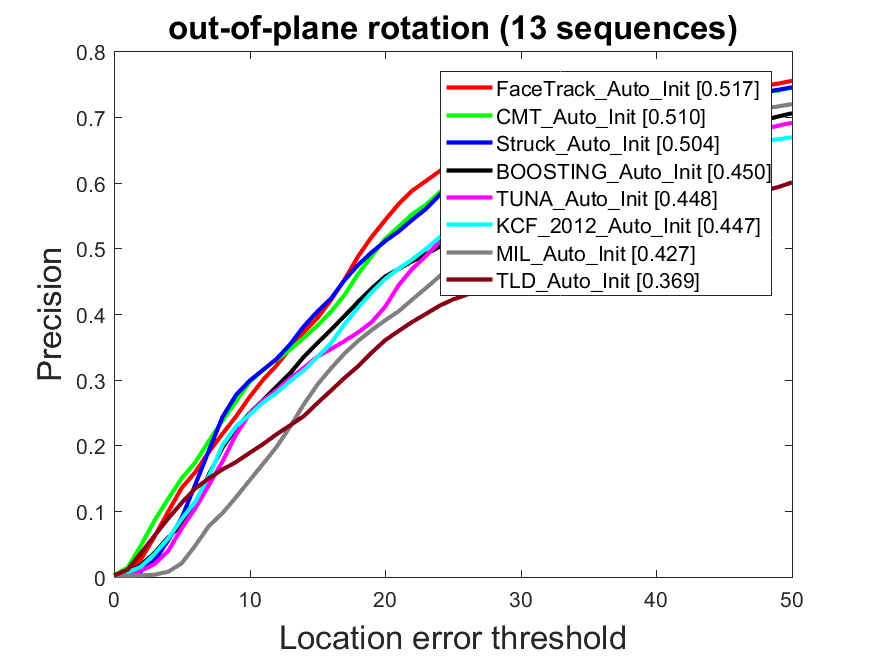}}
\subfloat[(d)]{\includegraphics[width=0.5\linewidth]{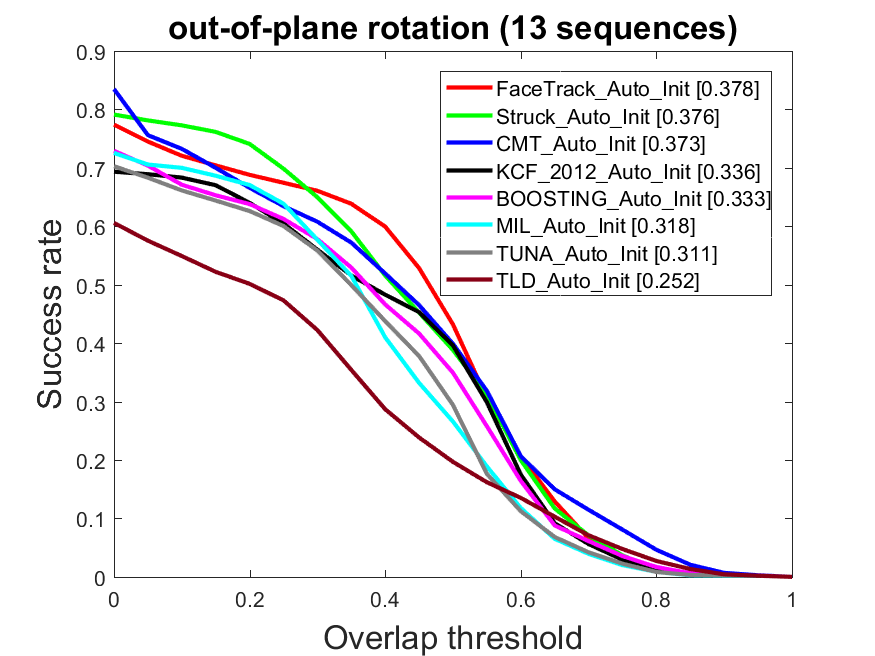}}\\
\subfloat[(e)]{\includegraphics[width=0.5\linewidth]{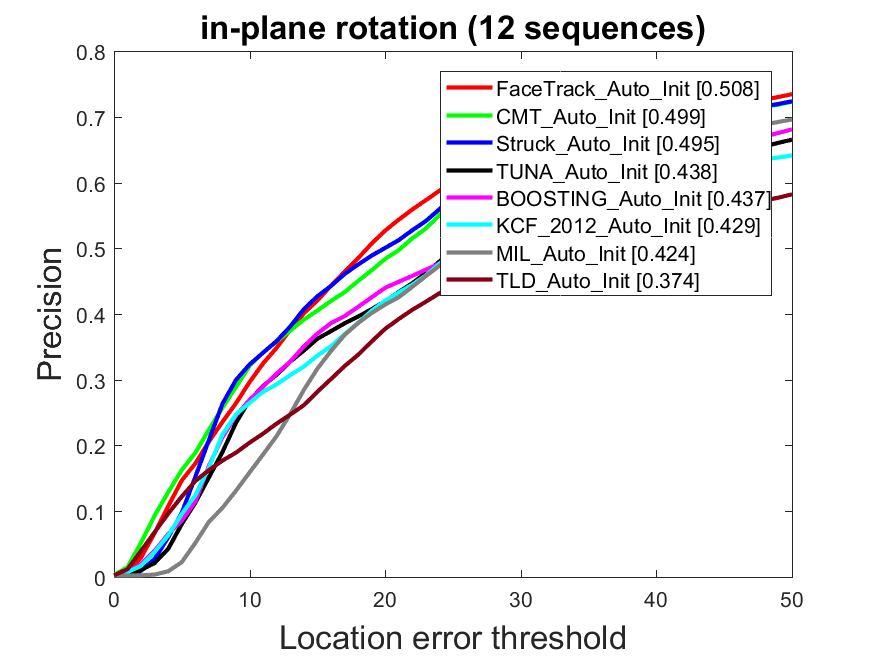}}
\subfloat[(f)]{\includegraphics[width=0.5\linewidth]{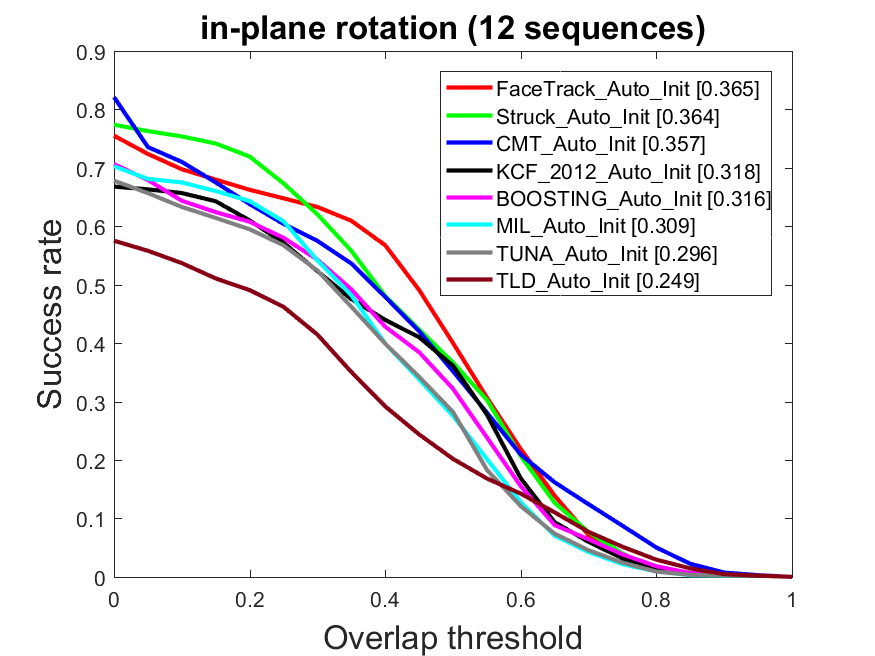}}
\caption{\footnotesize{FaceTrack performance on video attributes: (a) \& (b) scale variation, (c) \& (d) out-of-plane-rotation, (e) \& (f) in-plane-rotation (Best viewed when zoomed in.)}}
\label{fig:fig4}
\end{figure*}

\begin{figure*}[!htbp]
\vspace{-10em}
\captionsetup[subfigure]{labelformat=empty}
\centering
\subfloat[(a)]{\includegraphics[width=0.5\linewidth]{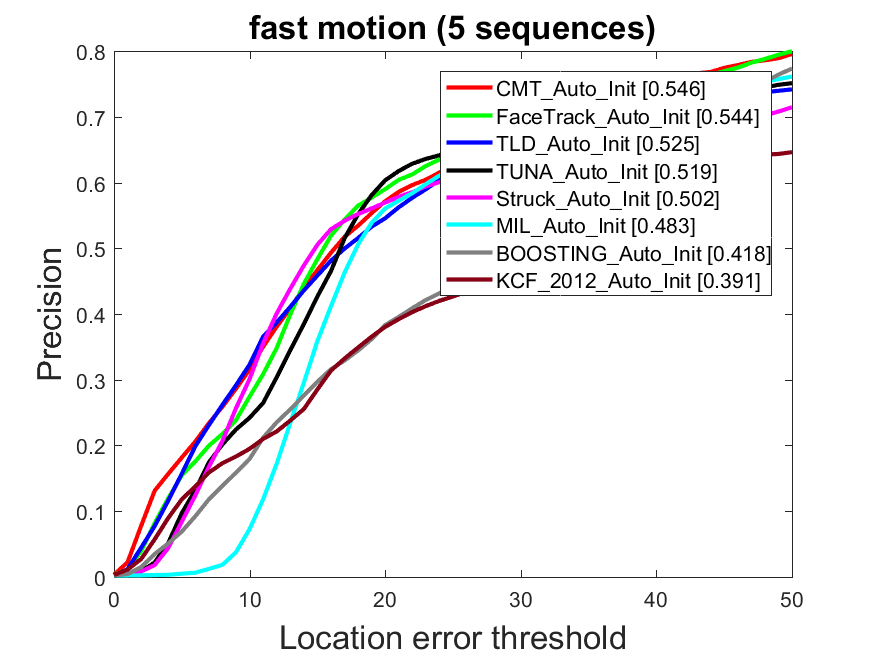}}
\subfloat[(b)]{\includegraphics[width=0.5\linewidth]{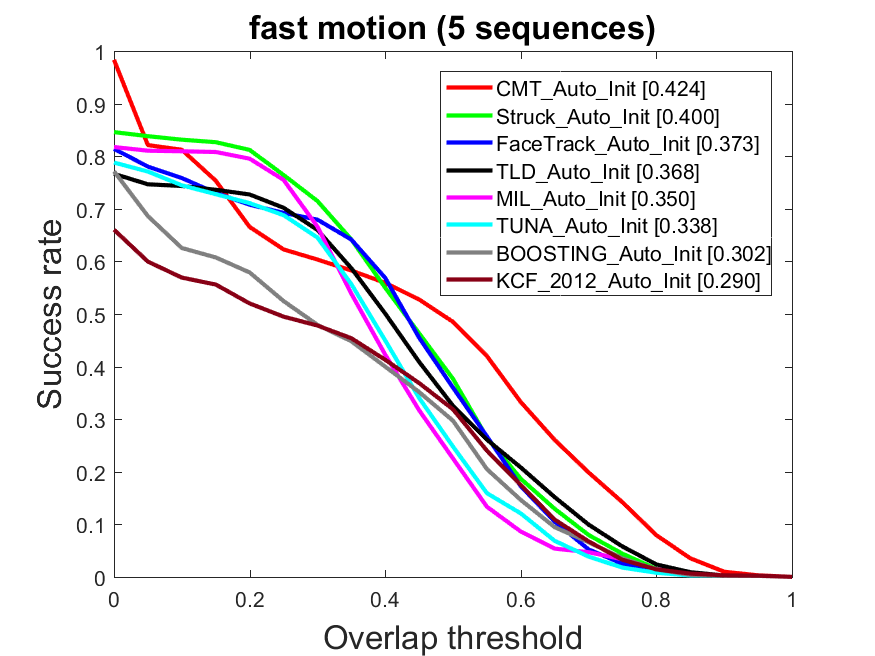}}\\
\subfloat[(c)]{\includegraphics[width=0.5\linewidth]{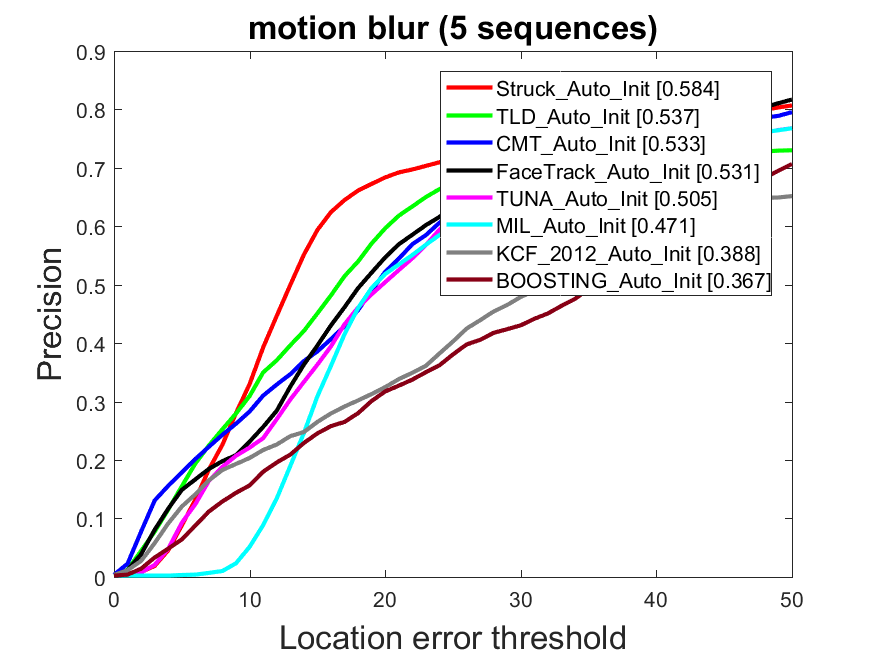}}
\subfloat[(d)]{\includegraphics[width=0.5\linewidth]{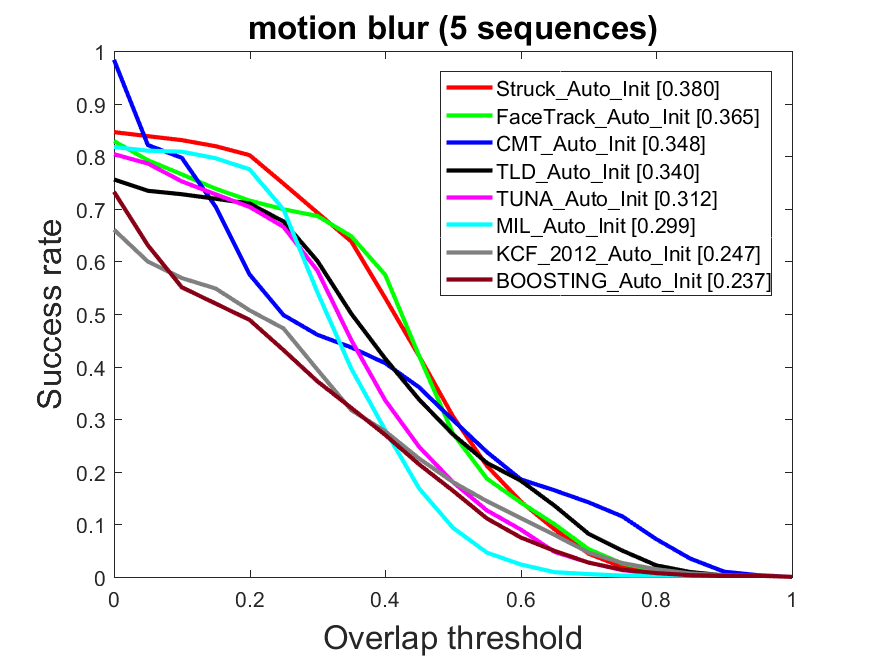}}\\
\subfloat[(e)]{\includegraphics[width=0.5\linewidth]{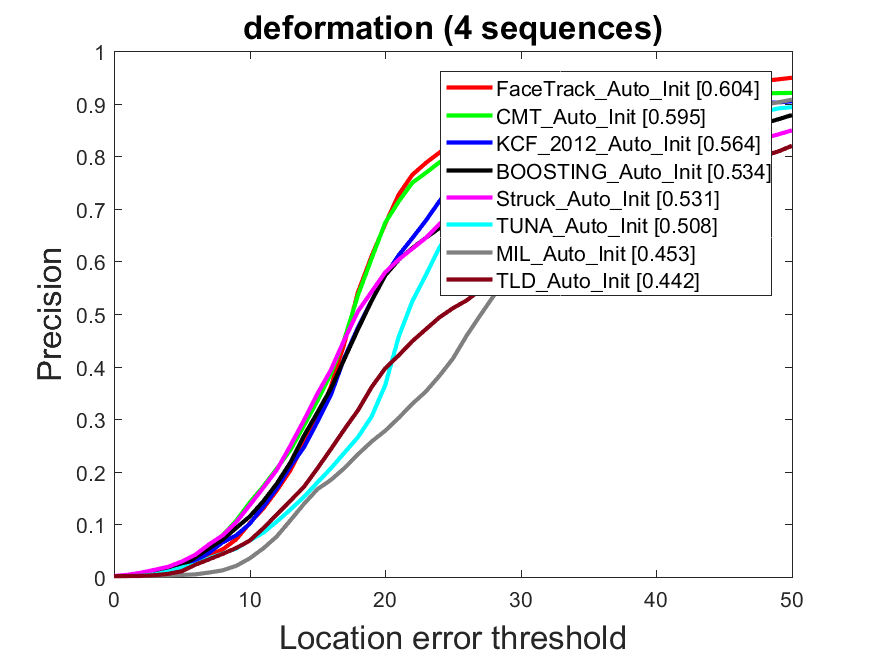}}
\subfloat[(f)]{\includegraphics[width=0.5\linewidth]{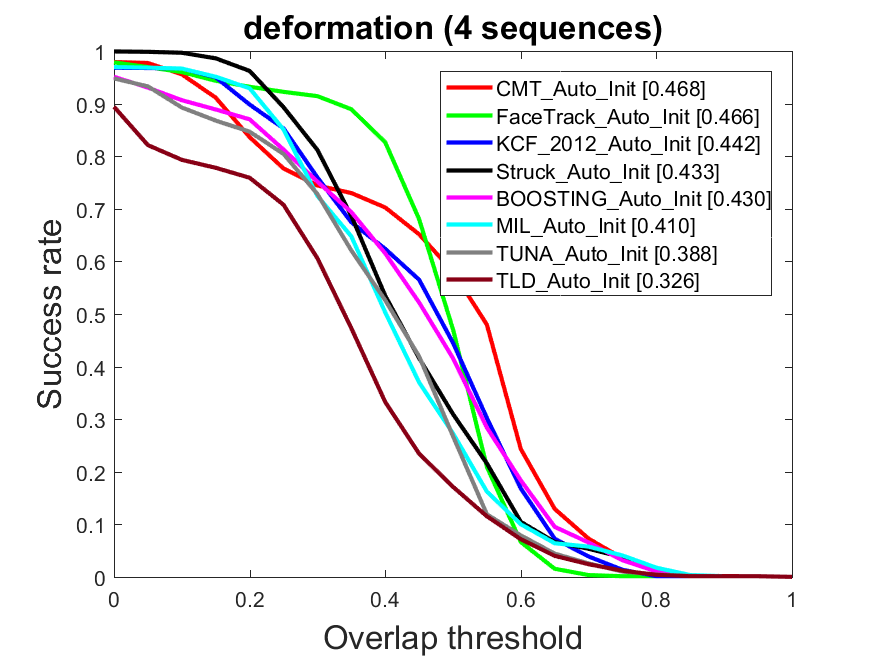}}\\
\caption{\footnotesize{FaceTrack performance on video attributes: (a) \& (b) fast motion, (c) \& (d) motion blur, (e) \& (f) deformation. (Best viewed when zoomed in.)}}
\label{fig:fig5}
\end{figure*}

\begin{figure*}[!htbp]
\vspace{-10em}
\captionsetup[subfigure]{labelformat=empty}
\centering
\subfloat[(o)]{\includegraphics[width=0.5\linewidth]{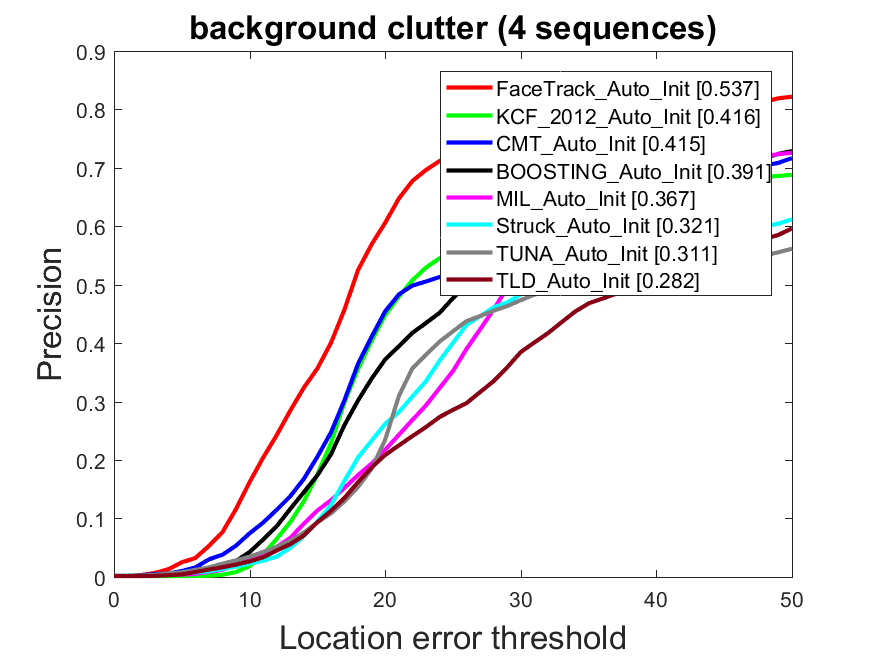}}
\subfloat[(p)]{\includegraphics[width=0.5\linewidth]{auto/background_clutter_error_OPE_AUC.png}}\\
\subfloat[(m)]{\includegraphics[width=0.5\linewidth]{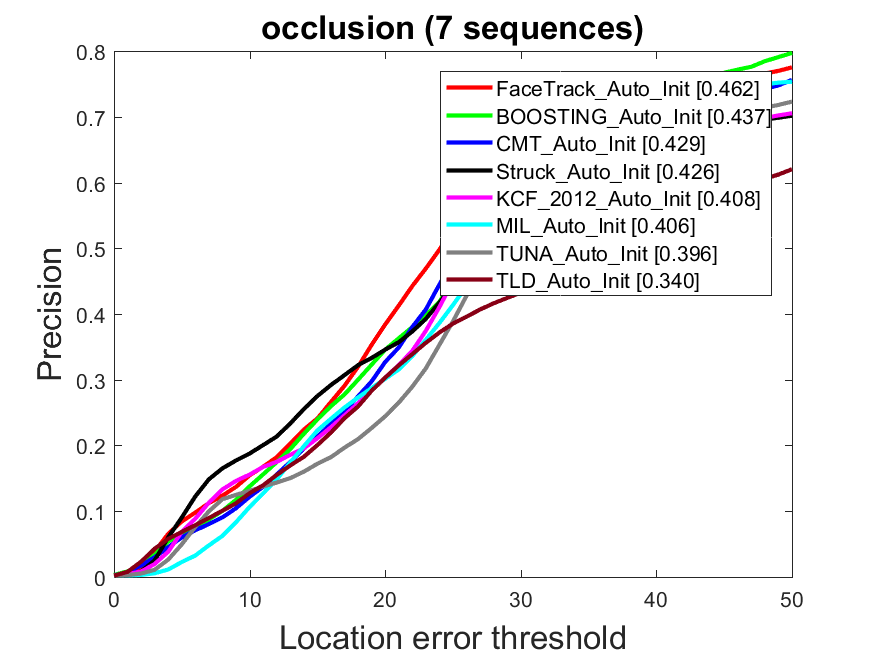}}
\subfloat[(n)]{\includegraphics[width=0.5\linewidth]{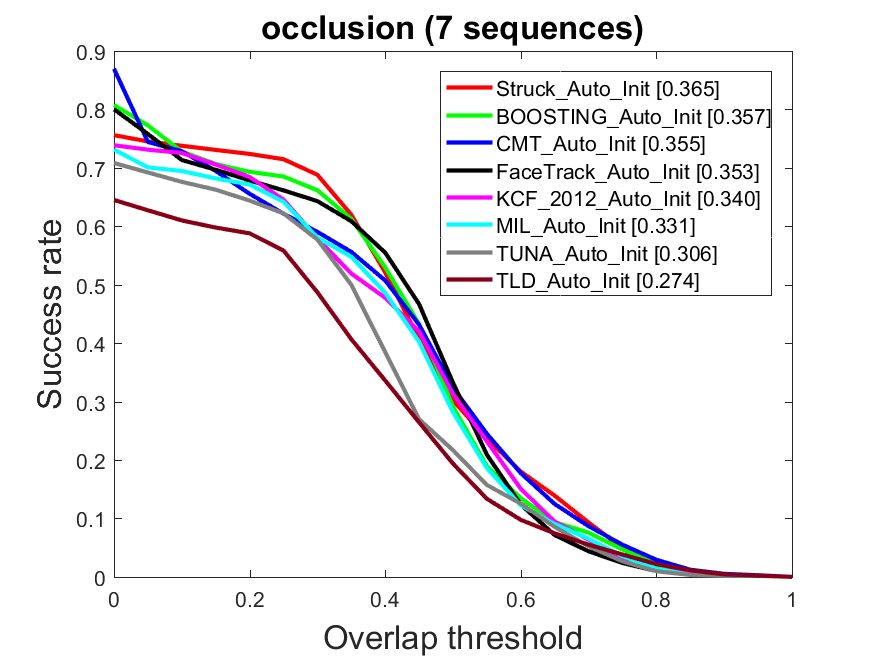}}\\
\subfloat[(q)]{\includegraphics[width=0.5\linewidth]{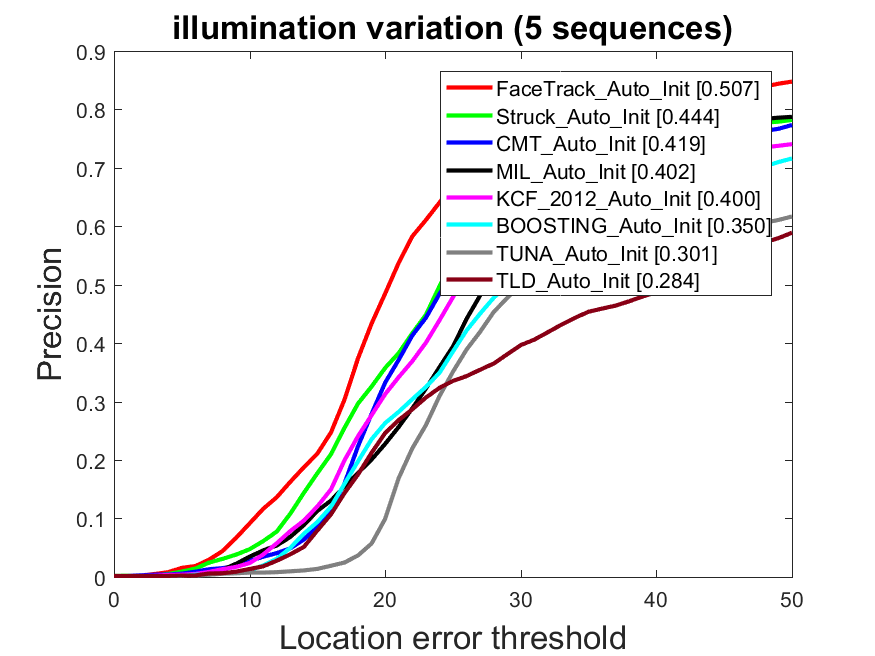}}
\subfloat[(r)]{\includegraphics[width=0.5\linewidth]{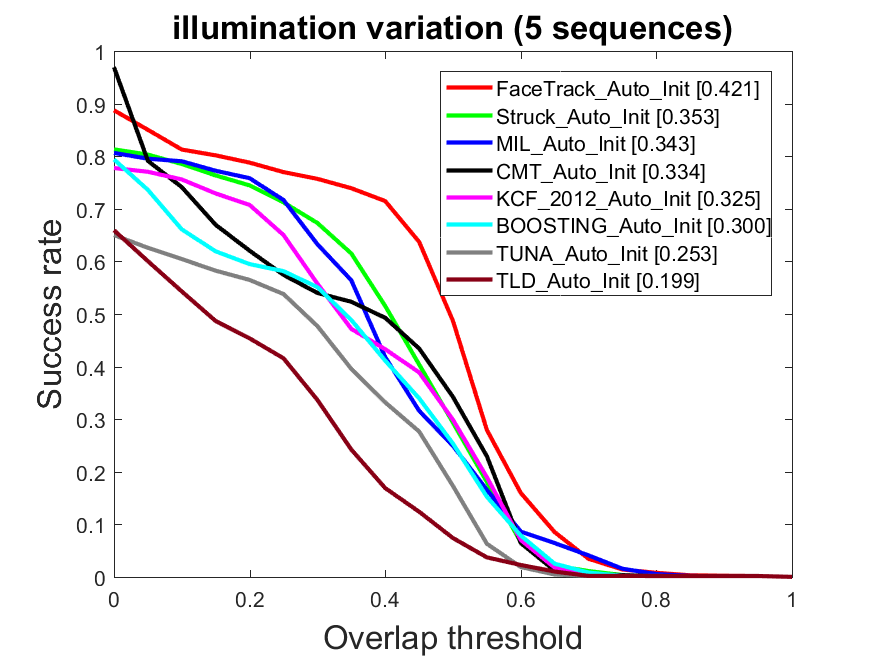}}
\caption{\footnotesize{FaceTrack performance on video attributes: (a) \& (b) background clutter, (c) \& (d) occlusion, (e) \& (f)  illumination variation. (Best viewed when zoomed in.)}}
\label{fig:fig6}
\end{figure*}

FaceTrack outperforms several state-of-the-art trackers by ranking first or second on almost all the tracking nuisances when initialized automatically (refer to Figure \ref{fig:fig4}, \ref{fig:fig5}, \ref{fig:fig6}). The following paragraph details the analysis.

\textbf{Scale variation and rotation}: Together with the keypoint scale adaptation strategy from \cite{TUNA}, and scale and aspect ratio adaptation from a face detector, the tracker performs well in tackling scale variation of the face, which is a common phenomena during object tracking. As long as the face remains partially or fully visible during in plane rotation and out-of-plane rotation, all the appearance models namely, GRM, ICM, and the BDM models contribute in face localization by maximizing the fusion score for all the face candidates. However, during out-of-plane rotation, GRM might be hidden and may not able to localize the face. On the other hand, because of the control and update strategy of our framework, the ICM and BDM templates get partially or fully updated (refer Algorithm \ref{algo:trackingUpdate}). Hence, during this time, ICM and BDM similarity score will dominate in maximizing the fusion score of face candidates.

\textbf{Fast motion and motion blur}: FaceTrack effectively deals with fast motion and motion blur during tracking by maximizing graph similarity in the whole frame. Further, having a face detector helps to find target during motion blur, since it does not suffer from the problem of drift due to its image independent searching principle (no spatio-temporal information is used).

\textbf{Background clutter}: The distinct appearance model GRM tackles the complex background and helps to identify the face during background clutter. During such scenarios, it becomes difficult to discriminate the face target from the background. But thanks to the L2-subspace based GRM appearance model that preserves the internal structural representation of the target face by assigning importance to the features that are memorized for long duration. Hence, the incremental learning of the model helps to capture the appearance representation and thus making it easier to track a face.

\textbf{Illumination variation and Occlusion}: When the target face undergoes severe illumination change in sequences, most of the methods tend to drift towards the cluttered background or cannot adapt to the scale change that occurs during this time. In addition, during this time and also during occlusion, the appearance of the target face changes drastically. Therefore, the GRM model is unable to localize the target face. Therefore, ICM and BDM, can be utilized for a short-term reference model for appearance matching. These models get updated frequently for short-term according to the update and control strategy in the proposed tracking framework. In addition, the face detector facilitates face localization and adaption of scale change during such drastic appearance change.

\textbf{Deformation}: The proposed tracker is able to handle object deformation very well in sequences. This is because during deformation, the $FDL$ associated with some of the keypoints in GRM may differ in length as the keypoints may get shifted from their original location. But the summing of the various kernel responses generated using multiple kernels in the response map, compensates for this error. Moreover, the face detector aids in reinitialization of the tracker in case the tracker drifts away from the target face. 

\textit{Failure cases}: The tracker may sometime loose track of a face in videos having drastic appearance change. In addition, it might be possible that the face detector is unable to detect the target face and output false positives. Thus, during this scenario, the similarity of face appearance cannot be established, due to which the face location might not get updated. Hence, during such a scenario the face might not get tracked. But, if a correct face detection for the target face can be obtained, then tracker will get re-initialized and will resume tracking. 

In summary, FaceTrack is able tackle the various tracking nuisances by utilizing the different components built in its algorithm. The next subsection presents an ablation analysis of FaceTrack.

\subsection{Ablation Analysis}
\begin{figure*}[!htbp]
\vspace{-1em}
\captionsetup[subfigure]{labelformat=empty}
\centering
\hspace{-3em}
\subfloat[(a)]{\includegraphics[width=0.5\linewidth]{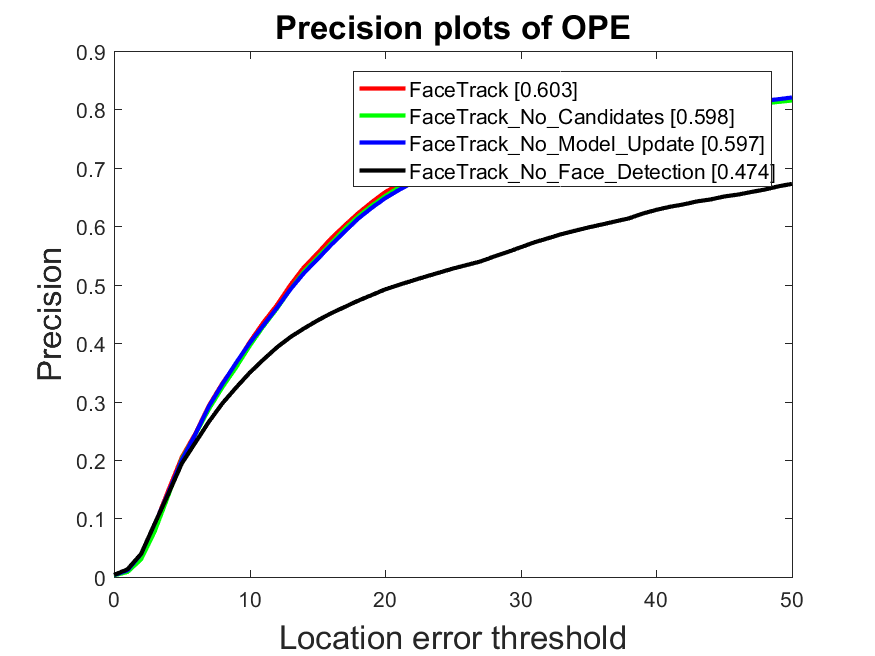}}
\subfloat[(b)]{\includegraphics[width=0.5\linewidth]{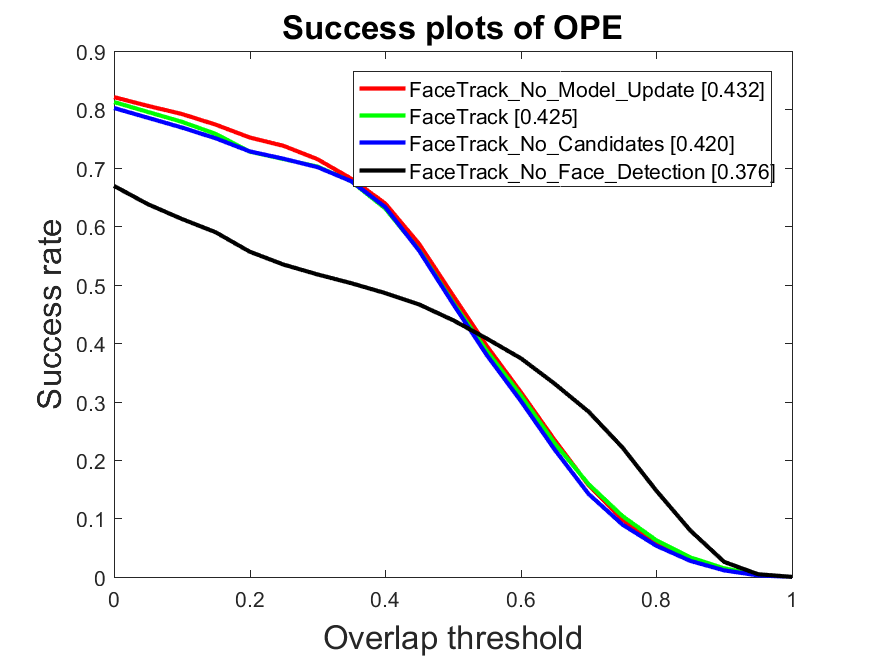}}
\caption{Ablation analysis of FaceTrack (a) Precision for One-Pass Evaluation (b) Success for One-Pass Evaluation. (Best viewed when zoomed in.)}
\label{fig:fig7}
\end{figure*}

To demonstrate the effectiveness of each component in the FaceTrack tracking framework, we eliminate in turn a component from it. For e.g. removing face detector, or removing the generation of face candidates or by not performing any appearance model updates for GRM, ICM and BDM respectively. It can be seen in Figure \ref{fig:fig7} that removing the face candidates from FaceTrack reduces its performance, which confirms our hypothesis that face candidates help in better face localization. Removing the face detector from FaceTrack results in performance loss indicating that face detector helps to tackle the drastic appearance changes during face tracking. By \textit{not} performing model update (both partial and full) of ICM and BDM models and not adding/deleting connections from GRM, but only doing updates using Equation \ref{eq:adapt} and control using Equation \ref{eq:theta} respectively, the performance of FaceTrack falls in precision but improves in success. This can be attributed to the fact that it is very challenging to perform correct appearance model update at all times during the online tracking process in the \textit{absence} of ground truth, which always involves a risk. On the other hand, if updates are not performed at all, then the face tracker might not be able to cope up with the changing appearance of face and eventually loose track. Thus, keeping all this in mind, it can be seen from the Figure \ref{fig:fig7} that even without updating ICM and BDM appearance models, FaceTrack is successfully able to track a face almost $60\%$ of the time, showcasing its robustness towards appearance change by adapting weights of keypoints present in the GRM model, and stability through its control strategy, tracking-by-detection using face detector, along with the weighted score-level fusion strategy for precise face target localization. Thus, all components play an important role in robust face tracking.

\subsection{Time Complexity Analysis}
By referring Algorithm \ref{algo:tracking}, it can be approximated that the time complexity of FaceTrack is $\approx$ $O(n^{2})$. FaceTrack estimates the face location as the face candidate having the highest fusion score. The similarity score of all the face candidates is obtained by maximizing the similarity of appearance models: GRM, ICM and BDM respectively. Please note that the matching of GRM model also accounts for keypoint extraction between two frames, matching keypoint descriptors between two frames and then finding the maximum in the kernel response map. The video sequences contain different frame resolution. FaceTrack runs with an average of 2 frames-per-second computed over 15 video sequences on an Intel Core i7 with a 3.40 GHz clock and 16GB RAM (single thread, with KeyPoint descriptor matching between two frames on NVIDIA GeForce GTX 560 graphic card).  

\section{Conclusion}\label{sec:Con}
In this paper, FaceTrack is proposed. It utilizes multiple appearance models for robust face tracking. The proposed multiple appearance models account for the \textit{temporal} (both long-term and short-term) appearance change of a face during tracking. FaceTrack jointly takes the advantage of the multiple appearance models by matching them effectively during different tracking scenarios to facilitate tracking. The incremental graph relational learning using the long-term update of face appearance features, helps to localize the face by finding an \textit{isomorphic} subgraph. The matched subgraph is approximated using multiple kernel functions in a kernel response map. The multiple kernels help to tackle face deformation and potential face tracking failures. In addition, the approximation also encodes error and eradicates its effect for precise face target location, by determining a non-linear decision boundary in the anisotropic kernel response map. In addition, the face detector helps to localize the face during drastic short-term appearance change and reinitialization of FaceTrack. Furthermore, for precise face location, face candidates are generated and the final face location is chosen as the candidate having the highest fusion score. Extensive experiments showcase the effectiveness of each component of the proposed face tracking framework for many tracking real-world unconstrained tracking nuisances in terms of accuracy, robustness, adaptiveness and tracking stability. In conclusion, it is essential that the face tracker should robustly adapt to appearance changes, and at the same time should output stable tracking results in spite of distractions which cannot be controlled in real-world scenarios.

\section{Acknowledgement}
This work was supported in part by FRQ-NT team grant \#167442 and by REPARTI (Regroupement pour l'\'etude des environnements partag\'es intelligents r\'epartis) FRQ-NT strategic cluster.

\section{References}

\bibliographystyle{elsarticle-num}
\bibliography{journal}

\begin{thebibliography}{10}
\expandafter\ifx\csname url\endcsname\relax
  \def\url#1{\texttt{#1}}\fi
\expandafter\ifx\csname urlprefix\endcsname\relax\def\urlprefix{URL }\fi
\expandafter\ifx\csname href\endcsname\relax
  \def\href#1#2{#2} \def\path#1{#1}\fi

\bibitem{saman}
S.~Bashbaghi, E.~Granger, R.~Sabourin, G.-A. Bilodeau, Robust watch-list
  screening using dynamic ensembles of svms based on multiple face
  representations, Machine Vision and Applications 28~(1) (2017) 219--241.

\bibitem{Liao:2016:FAU:2914182.2914310}
S.~Liao, A.~K. Jain, S.~Z. Li, A fast and accurate unconstrained face detector,
  IEEE Trans. Pattern Anal. Mach. Intell. 38~(2) (2016) 211--223.

\bibitem{TLD}
Z.~Kalal, K.~Mikolajczyk, J.~Matas, Tracking-learning-detection, IEEE Trans.
  Pattern Anal. Mach. Intell. 34~(7) (2012) 1409--1422.

\bibitem{boosting}
H.~Grabner, M.~Grabner, H.~Bischof, Real-time tracking via on-line boosting,
  in: Proc. BMVC, 2006, pp. 6.1--6.10.

\bibitem{Struck}
S.~Hare, A.~Saffari, P.~H.~S. Torr, Struck: Structured output tracking with
  kernels (2011) 263--270.

\bibitem{MIL}
X.~Mei, H.~Ling, Robust visual tracking using l1 minimization, in: {IEEE} 12th
  International Conference on Computer Vision, {ICCV} 2009, Kyoto, Japan,
  September 27 - October 4, 2009, 2009, pp. 1436--1443.

\bibitem{KCF}
J.~Henriques, R.~Caseiro, P.~Martins, J.~Batista, Exploiting the circulant
  structure of tracking-by-detection with kernels, in: Proceedings of the 12th
  European Conference on Computer Vision - Volume Part IV, ECCV'12,
  Springer-Verlag, Berlin, Heidelberg, 2012, pp. 702--715.

\bibitem{IVT}
D.~A. Ross, J.~Lim, R.-S. Lin, M.-H. Yang, Incremental learning for robust
  visual tracking, Int. J. Comput. Vision 77~(1-3) (2008) 125--141.

\bibitem{6619151}
D.~Wang, H.~Lu, M.~H. Yang, Least soft-threshold squares tracking, in: 2013
  IEEE Conference on Computer Vision and Pattern Recognition, 2013, pp.
  2371--2378.

\bibitem{6926840}
J.~Lu, G.~Wang, W.~Deng, K.~Jia, Reconstruction-based metric learning for
  unconstrained face verification, IEEE Transactions on Information Forensics
  and Security 10~(1) (2015) 79--89.

\bibitem{Ehsan}
E.~Elhamifar, R.~Vidal, Sparse subspace clustering: Algorithm, theory, and
  applications, IEEE Transactions on Pattern Analysis and Machine Intelligence
  35~(11) (2013) 2765--2781.

\bibitem{aam}
S.~Salti, A.~Cavallaro, L.~Di~Stefano, Adaptive appearance modeling for video
  tracking: Survey and evaluation, Trans. Img. Proc. 21~(10) (2012) 4334--4348.

\bibitem{Dewan}
M.~A.~A. Dewan, E.~Granger, G.-L. Marcialis, R.~Sabourin, F.~Roli, Adaptive
  appearance model tracking for still-to-video face recognition, Pattern
  Recognition 49 (2016) 129 -- 151.

\bibitem{TUNA}
T.~Chakravorty, G.~Bilodeau, E.~Granger, Tracking using numerous anchor points,
  CoRR abs/1702.02012.

\bibitem{CTSE}
T.~Chakravorty, G.~A. Bilodeau, E.~Granger, Contextual object tracker with
  structure encoding, in: 2015 IEEE International Conference on Image
  Processing (ICIP), 2015, pp. 4937--4941.

\bibitem{CMT}
G.~Nebehay, R.~Pflugfelder, Consensus-based matching and tracking of keypoints
  for object tracking, in: IEEE Winter Conference on Applications of Computer
  Vision, 2014, pp. 862--869.

\bibitem{Shi:2000:NCI:351581.351611}
J.~Shi, J.~Malik, Normalized cuts and image segmentation, IEEE Trans. Pattern
  Anal. Mach. Intell. 22~(8) (2000) 888--905.

\bibitem{Holder:2003:GRL:959242.959254}
L.~B. Holder, D.~J. Cook, Graph-based relational learning: Current and future
  directions, SIGKDD Explor. Newsl. 5~(1) (2003) 90--93.

\bibitem{Belkin:2003:LED:795523.795528}
M.~Belkin, P.~Niyogi, Laplacian eigenmaps for dimensionality reduction and data
  representation, Neural Comput. 15~(6) (2003) 1373--1396.

\bibitem{Lowe:1999:ORL:850924.851523}
D.~G. Lowe, Object recognition from local scale-invariant features, in:
  Proceedings of the International Conference on Computer Vision-Volume 2 -
  Volume 2, ICCV '99, IEEE Computer Society, Washington, DC, USA, 1999.

\bibitem{lbsp}
P.~L. St-Charles, G.~A. Bilodeau, Improving background subtraction using local
  binary similarity patterns, in: IEEE Winter Conference on Applications of
  Computer Vision, 2014, pp. 509--515.

\bibitem{ootb}
Y.~Wu, J.~Lim, M.~H. Yang, Online object tracking: A benchmark, in: 2013 IEEE
  Conference on Computer Vision and Pattern Recognition, 2013, pp. 2411--2418.

\end{thebibliography}

\end{document}